\documentclass{article}

\usepackage{graphicx}
\usepackage{subfigure}
\usepackage{booktabs} 

\usepackage{amsfonts, amsmath, amsthm}
\usepackage[inline]{enumitem}
\usepackage{multirow}
\usepackage{algorithm}
\usepackage{algorithmic}
\usepackage{authblk}
\newcommand{\citet}{\cite}
\newcommand{\citealt}{\cite}

\DeclareMathOperator*{\expect}{\mathbb{E}}

\renewcommand{\algorithmicrequire}{\textbf{Input:}}
\renewcommand{\algorithmicensure}{\textbf{Output:}}

\newtheorem{theorem}{Theorem}

\newif\iflong
\newif\ifthesis

\longtrue

\title{Gaussian Process Classification \\ with Privileged Information \\
	by Soft-to-Hard Labeling Transfer}

\author[1]{Ryosuke Kamesawa}
\author[1,2]{Issei Sato}
\author[2,1]{Masashi Sugiyama}
\affil[1]{The University of Tokyo, Japan}
\affil[2]{RIKEN, Japan}
\affil[ ]{\texttt{\{kamesawa@ms., sato@, sugi@\}k.u-tokyo.ac.jp}}

\begin{document}

\maketitle

\begin{abstract}
\textit{Learning using privileged information} is an attractive problem setting
that helps many learning scenarios in the real world.
A state-of-the-art method of Gaussian process classification (GPC)
with privileged information is GPC+,
which incorporates privileged information into a noise term of the likelihood.
A drawback of GPC+ is that
it requires numerical quadrature
to calculate the posterior distribution of the latent function,
which is extremely time-consuming.
To overcome this limitation,
we propose a novel classification method with privileged information
based on Gaussian processes, called
``soft-label-transferred Gaussian process (SLT-GP).''
Our basic idea is that we construct another learning task of
predicting soft labels (continuous values)
obtained from privileged information and
we perform transfer learning from this task to
the target task of predicting hard labels.
We derive a PAC-Bayesian bound of our proposed method,
which justifies optimizing hyperparameters by the empirical Bayes method.
We also experimentally show the usefulness of our proposed method
compared with GPC and GPC+.
\end{abstract}

\section{Introduction}

\textit{Learning using privileged information}~\cite{vapnik2009NN} is
an attractive problem setting that helps many learning scenarios
in the real world~\cite{vapnik2015JMLR}.
The first method that can solve the problem of learning using privileged information
is the support vector machine plus (SVM+)~\cite{vapnik2009NN}.
SVM+ uses privileged information to predict the optimal slack variables in SVM\@.
It has been proven that the convergence of the generalization error of SVM+ is
faster than that of standard SVM\@.
However, the accuracy of SVM+ has been empirically reported not to be
so much better than that of standard SVM~\cite{sharmanska2013ICCV, hernandez2014NIPS}.

Gaussian process classification (GPC)~\cite{williams1998IEEE},
which enables us to model the uncertainty of prediction, has been extended
to GPC+~\cite{hernandez2014NIPS} so as to use privileged information.
GPC+ incorporates privileged information into a noise term of the likelihood.
However, the formulation of GPC+ requires numerical quadrature
to calculate the posterior distribution.
Since the number of numerical quadratures performed in GPC+ is proportional to
the number of training examples,
computation of GPC+ takes a long time.
Moreover, it is hard to theoretically analyze
the expected classification risk in a PAC-Bayes way~\cite{mcallester1998COLT}
because the negative log likelihood loss function
cannot be defined from only the input and the output in the case of GPC+.

\begin{table*}
	\centering
	\caption{Comparison between three classification methods
using Gaussian processes.
The likelihood function is modified in GPC+ so as to incorporate privileged information
while the likelihood function is the same as GPC in the case of SLT-GP\@.
In SLT-GP, the prior can be regarded to be modified by
soft labels.}\label{tab:comparison}
\makebox[\linewidth]{
	\begin{tabular}{cccc}
		\toprule
		& GPC & GPC+ & SLT-GP \\
		& \cite{kuss2005JMLR} & \cite{hernandez2014NIPS} & (This work) \\
		\midrule
		\multirow{2}{*}{Prior} &
		\multirow{2}{*}{$f \sim \mathcal{GP}(0, k)$} &
		$f \sim \mathcal{GP}(0, k)$ &
		\multirow{2}{*}{$f_T \sim \mathcal{GP}(m_{X,\mathbf{s}}, k_X)$} \\
		& & $g \sim \mathcal{GP}(m^*, k^*)$ & \\
		Likelihood &
		$p(y|f, x) = \Phi(yf(x))$ &
		$p(y|f, g, x, x^*) = \Phi\left(\frac{yf(x)}{\sqrt{\exp(g(x^*))}}\right)$ &
		$p(y|f_T, x) = \Phi\left(yf_T(x))\right)$ \\
		Update in EP & analytic & with numerical quadrature & analytic \\
		PAC-Bayes bound & $\surd$~\cite{seeger2002JMLR} & n/a &
		$\surd$ (Section~\ref{sec:bound}) \\
		\bottomrule
	\end{tabular} }
\end{table*}

To overcome the above limitations of GPC+,
we propose a novel classification method with privileged information
based on Gaussian processes.
We formulate the classification problem with privileged information
as a soft-label-transferred Gaussian processes (SLT-GP).
We assume that there is a relationship between two tasks:
\begin{enumerate*}[label={(\alph*)}]
\item predicting hard labels and
\item predicting soft labels.
\end{enumerate*}
Hard labels are binary values provided in a training dataset,
while soft labels are continuous values extracted from privileged information
similarly to the framework of
\textit{generalized distillation}~\cite{lopez2016ICLR}.
We then formulate the classification problem with privileged information as
transfer learning from the soft-labeling task to the hard-labeling task.
In our formulation, we can derive an efficient expectation propagation algorithm
to approximate the posterior distribution.
Also, we can analyze the expected risk of our proposed method
by using the PAC-Bayesian theorem~\cite{germain2016NIPS}
because it can be regarded that the prior distribution is modified
from that of GPC in our formulation
while the likelihood function is modified from that of GPC in the formulation of GPC+.
Our theoretical analysis justifies the optimization of
hyperparameters by the empirical Bayes method.
Table~\ref{tab:comparison} summarizes the relationship
between standard GPC, GPC+, and our SLT-GP method proposed in this paper.
Finally, we experimentally demonstrate the usefulness of
the SLT-GP method.

\section{Proposed formulation}

In this section, we provide background information and
the formulation for the model of the proposed method.
The learning algorithm for the proposed method
is described in the next section.

\subsection{Preliminary}

We prepare several notations for datasets.
A training dataset consists of training input data $x_i \in \mathcal{X}$,
target labels $y_i \in \{+1, -1\}$,
and privileged information $x_i^* \in \mathcal{X}^*$,
where $\mathcal{X}$ is the input space
and $\mathcal{X}^*$ is the space of privileged information.
Note that since our method is based on Gaussian processes and it works with
only covariance functions (and mean functions),
$\mathcal{X}$ and $\mathcal{X}^*$ can be any spaces
as long as covariance functions are defined on those spaces.

The overall dataset with $n$ training examples
is denoted by $D = \{(x_i, y_i, x_i^*)\}_{i=1}^n$.
For brevity, we introduce some additional notations for the training dataset,
$X = (x_1, \ldots, x_n) \in \mathcal{X}^{n}$,
$X^* = (x_1^*, \ldots, x_n^*) \in (\mathcal{X}^*)^n$, and
$\mathbf{y} = (y_1, \ldots, y_n)^\top \in \mathbb{R}^n$,
where $\top$ denotes the transpose of a vector and a matrix.

\subsection{Soft labels}

Our method uses soft labels extracted from
GPC with privileged information.
Soft labels are more informative than hard labels
because soft labels have continuous values
while hard labels have only sign information.
Soft labels represents the degree of confidence of belonging to each class.
To extract soft labels from the privileged information,
we use standard GPC\@.
The Gaussian process over the space of privileged information is defined by
$f^* \sim \mathcal{GP}(0, k^*)$,
where $\mathcal{GP}(m, k)$ denotes the Gaussian process with mean function $m$ and
covariance function $k$, $0$ is the zero function, and
$k^*: \mathcal{X}^* \times \mathcal{X}^* \rightarrow \mathbb{R}$ is
a covariance function over the space of privileged information $\mathcal{X}^*$.
With an appropriate likelihood for classification, e.g.\
the probit model $p(y=1|f(x)) = \Phi(f(x))$, the posterior distribution
$p(f^* | X^*, y)$ can be approximated,
where $\Phi(z)$ is the cumulative distribution function
of the standard normal distribution.
Especially, the posterior distribution of
$\mathbf{f}^* = (f^*(x_1^*), \ldots, f^*(x_n^*))^\top$ can be approximated as
a multivariate normal distribution,
\begin{align}
	p(\mathbf{f}^* | X^*, y) \simeq q(\mathbf{f}^*) =  \mathcal{N}(\mathbf{f}^* | \boldsymbol{\mu}^*, \Sigma^*),
\end{align}
where $\boldsymbol{\mu}^*$ is the approximated posterior mean vector and
$\Sigma$ is the posterior covariance matrix.
The posterior mean is treated as soft labels, $s_i = \mu_i^*$,
and equivalently denoted by
$\mathbf{s} = (s_1, \ldots, s_n)^\top = \boldsymbol{\mu}^*$.

\subsection{Soft-to-hard labeling transfer}

Once the soft labels are obtained,
we can consider two tasks: (a) the task predicting the hard label $y_i$
from the input $x_i$ and
(b) the task predicting the soft label $s_i$ from the input $x_i$.
By using the soft labels,
we do not need to care about the privileged information itself
since it is expected that the soft labels have enough information
to predict the target labels as well as the privileged information.
These two tasks are highly related when the privileged information $x_i^*$ has
useful information about the input $x_i$ to predict the target label $y_i$,
that is, the soft label $s_i$ represents
the appropriate confidence of being positive or negative for the input $x_i$.
Furthermore, it might be easier to predict the target label
thorough predicting soft labels than predicting hard labels directly
since soft labels have richer information than hard labels.
This is similar to the idea of SVM+,
which conceptually predicts the slack variables in the SVM
using the privileged information~\cite{vapnik2015JMLR}.
However, there is no guarantee that the privileged information has
the appropriate information about the data in real-world datasets.
In such cases, making a prediction only from the soft labels causes
severe performance deterioration.

To cope with this problem,
we propose using transfer learning with Gaussian processes.
The target task is predicting binary hard labels and
the source task is predicting soft labels.
Thus, the source task is a regression problem and
the target task is a classification problem.
We should note that the domains of these tasks are the same, $\mathcal{X}$,
and the training input data $X$ are also the same for both tasks.
A similar formulation is explored
in multi-task setting~\cite{bonilla2007NIPS},
in transfer setting where
both source and target tasks are regression problems~\cite{cao2010AAAI},
and in multi-output setting~\cite{alvarez2010AISTATS}.

In our proposed method,
both the source and target tasks have the same Gaussian process
as a prior distribution over latent functions.
Both Gaussian process regression and classification assume that
a Gaussian process prior has the covariance function over the input space,
$k: \mathcal{X} \times \mathcal{X} \rightarrow \mathbb{R}$.
The prior distributions over the latent functions of the source task $f^{(S)}$ and
the target task $f^{(T)}$ are common:
$f^{(S)}, f^{(T)} \sim \mathcal{GP}(0, k)$.
To define a Gaussian process over multiple tasks,
the inter-task covariance should be defined. 
The covariance function between the source task and the target task is
defined as follows:
\begin{align}
	\mathbb{E}[(f^{(S)}(x), f^{(T)}(x'))] = \rho k(x, x'),
\end{align}
where $\rho \in [0, 1]$ is a \textit{task similarity} between the source and target tasks.
This task similarity represents the relationship between hard labels and soft labels
extracted from the privileged information.
If the task similarity is close to 1, the latent functions of both tasks have
close outcomes.
Thus the information from soft labels are fully incorporated in the target task.
If the task similarity is 0, there is no similarity
between the outcomes of the source and target latent functions.
Therefore, the latent functions $f^{(S)}$ and $f^{(T)}$ are independent
and the soft labels are ignored.
The task similarity parameter can be estimated by the empirical Bayes method
as explained later.

The difference between the source task and the target task is reduced to the
difference of their likelihood functions.
On the one hand, the source task, predicting soft labels, is essentially
a regression problem.
Thus, the likelihood for the source task is modeled by a Gaussian likelihood function,
$p(s_i | f_i^{(S)}) = \mathcal{N}(s_i | f_i^{(S)}, 1)$,
where $f_i^{(S)} = f^{(S)}(x_i)$.
On the other hand, the target task is a classification problem.
Thus, the likelihood for the target task is modeled by a Bernoulli likelihood function,
$p(y_i = 1 | f_i^{(T)}) = \Phi(f_i^{(T)})$,
where $f_i^{(T)} = f^{(T)}(x_i)$.
Since the likelihood terms in the source and target domains are different,
we need to devise a learning algorithm for this transfer formulation.

\section{Learning algorithm}\label{sec:algorithm}

Since the likelihood of the target tasks is non-Gaussian,
the posterior distribution of $f^{(T)}$ is analytically intractable.
Therefore, we need to approximate the posterior distribution.
We use expectation propagation (EP)~\cite{minka2001UAI} to approximate
the posterior Gaussian process.
The difference between EP
of the standard GPC and that of our method is the likelihood term.
The likelihood term for GPC is the Bernoulli distribution
while the likelihood term for our method includes both the
Bernoulli and Gaussian likelihoods.
We propose an EP method for a mixture of multiple likelihoods
to estimate the posterior distribution in our method.

\subsection{Expectation propagation}\label{subsec:ep_slt}

From Bayes' theorem, the posterior distribution of the latent functions
$f^{(T)}$ and $f^{(S)}$ are derived as follows:
\begin{multline}
	p(f^{(T)}, f^{(S)} | X, \mathbf{y}, \mathbf{s}) \\
	= \frac{p(\mathbf{y} | f^{(T)}, X) p(\mathbf{s} | f^{(S)}, X) p(f^{(T)}, f^{(S)})}
	{p(\mathbf{y}, \mathbf{s} | X)},
\end{multline}
where $p(\mathbf{y}, \mathbf{s} | X)$ is the marginal likelihood
integrating out $f^{(S)}$ and $f^{(T)}$.
The same equation holds for the latent function values at the finite training points
$\mathbf{f}_T = (f_1^{(T)}, \ldots, f_n^{(T)})^\top$ and
$\mathbf{f}_S = (f_1^{(S)}, \ldots, f_n^{(S)})^\top$ as follows:
\begin{gather}
	p(\mathbf{f}_T, \mathbf{f}_S | X, \mathbf{y}, \mathbf{s}) =
	\frac{p(\mathbf{y}|\mathbf{f}_T) p(\mathbf{s}|\mathbf{f}_S)
		p(\mathbf{f}_T, \mathbf{f}_S)
	}{p(\mathbf{y}, \mathbf{s} | X)
	} \nonumber \\
	= \frac{\prod_{i=1}^n [p(y_i | f_i^{(T)}) p(s_i | f_i^{(S)})]
		p(\mathbf{f}_T, \mathbf{f}_S)
	}{p(\mathbf{y}, \mathbf{s} | X)},
\end{gather}
where
\begin{align}
	p(\mathbf{f}_T, \mathbf{f}_S) &=
	\mathcal{N}\left(
		\begin{bmatrix}
			\mathbf{f}_T \\ \mathbf{f}_S
		\end{bmatrix} \middle|
		\boldsymbol{0}, K
	\right), \\
	K &= \begin{pmatrix}
		1 & \rho \\
		\rho & 1
	\end{pmatrix} \otimes K_X, \\
	K_X &= \left(k(x_i, x_j)\right)_{i,j},
\end{align}
and $\otimes$ denotes the Kronecker product.
This is derived from the property of the Gaussian process we define.

While all the likelihood terms are approximated in GPC,
only the likelihood for the target tasks is approximated in our method.
To approximate the posterior distribution
$p(\mathbf{f}_T, \mathbf{f}_S | X, \mathbf{y}, \mathbf{s})$,
the non-Gaussian likelihood $p(\mathbf{y} | \mathbf{f}_T)$ is
approximated by an unnormalized Gaussian function of $\mathbf{f}_T$
called ``site approximation,''
\begin{align}
	p(y_i | f_i^{(T)}) \simeq t_i^{(T)}(f_i^{(T)})
	:= \widetilde{Z}_i \mathcal{N}(f_i^{(T)} |
	\widetilde{\mu}_i, \widetilde{\sigma}_i),
\end{align}
where $\widetilde{Z}_i$ is the normalization term.
This is also denoted by natural parameters of the normal distribution.
The moment parameters $\widetilde{\mu}_i$ and $\widetilde{\sigma}^2$ are
converted into natural parameters $\widetilde{\nu}_i$ and
$\widetilde{\tau}$ as follows:
\begin{align}
	\widetilde{\nu}_i = \frac{\widetilde{\mu}_i}{\widetilde{\sigma}^2}_i, \quad
	\widetilde{\tau}_i = \frac{1}{\widetilde{\sigma}^2}_i.
\end{align}
For the Gaussian likelihood terms, no approximation is required,
\begin{align}
	p(s_i | f_i^{(S)}) = t_i^{(S)}(f_i^{(S)})
	:= \mathcal{N}(f_i^{(S)} | s_i, 1).
\end{align}
Once the likelihood terms are approximated by the site approximations
$t_i^{(T)}(f_i^{(T)})$, the approximated posterior distribution
$q(\mathbf{f}) \simeq p(\mathbf{f}| X, \mathbf{y}, \mathbf{s})$
over both the source and target
$\mathbf{f} = \begin{bmatrix}\mathbf{f}^{(T)} \\ \mathbf{f}^{(S)})\end{bmatrix}$
can be computed,
\begin{align}\label{eq:slt_post_update}
	q(\mathbf{f}) &= \mathcal{N}(\mathbf{f} | \boldsymbol{\mu}, \Sigma), \\
	\boldsymbol{\mu} &= \Sigma \widetilde{\Sigma}^{-1} \widetilde{\boldsymbol{\mu}}
	= \Sigma \widetilde{\boldsymbol{\nu}}, \\
	\Sigma &= (K^{-1} + \widetilde{\Sigma}^{-1})^{-1} \nonumber \\
	&= K - K \widetilde{T}^\frac{1}{2} B^{-1} \widetilde{T}^\frac{1}{2} K,
\end{align}
where 
\begin{align}
	\widetilde{\boldsymbol{\mu}} &= (
		\widetilde{\mu}_1, \ldots, \widetilde{\mu}_n,
	s_1, \ldots, s_n)^\top \in \mathbb{R}^{2n}, \\
	\widetilde{\Sigma} &= \text{diag}(
		\widetilde{\sigma}_1^2, \ldots, \widetilde{\sigma}_n^2,
	1, \ldots, 1) \in \mathbb{R}^{2n\times 2n}, \\
	\widetilde{\boldsymbol{\nu}} &=
	\left(\widetilde{\nu}_1, \ldots, \widetilde{\nu}_n,
	s_1, \ldots, s_n\right)^\top
	\in \mathbb{R}^{2n},\\
	\widetilde{T} &= \text{diag}(\widetilde{\tau}_1, \ldots, \widetilde{\tau}_n,
	1, \ldots, 1)
	\in \mathbb{R}^{2n\times 2n}, \\
	B &= I + \widetilde{T}^\frac{1}{2} K \widetilde{T}^\frac{1}{2}.
\end{align}
The site approximations $t_i^{(T)}(f_i^{(T)})\;(i=1,\ldots,n)$ are updated iteratively
in the same way as the standard GPC\@.

\subsection{Inference}

After estimating the site approximation $t_i^{(T)}(f_i^{(T)})$,
we can infer the predictive distribution for an unknown input $\hat{x}$.
Since the likelihood terms are approximated by unnormalized Gaussian functions,
the predictive distribution of the latent function value
can be calculated in the same way as the regression case:
\begin{align}
	p(f(\hat{x}) | \hat{x}, X, \mathbf{y}, \mathbf{s})
	&\simeq \mathcal{N}(f(\hat{x}) | \hat{\mu}, \hat{\sigma}^2), \\
	\hat{\mu} &= \widehat{\mathbf{k}}^\top
	(K + \widetilde{T}^{-1})^{-1} \widetilde{\mu} \nonumber \\
	&= \widehat{\mathbf{k}}^\top
	(I + \widetilde{T}^{\frac{1}{2}} B^{-1}
	\widetilde{T}^{\frac{1}{2}} K)^{-1}
	\widetilde{\boldsymbol{\nu}}, \\
	\hat{\sigma}^2 &= k(\hat{x}, \hat{x})
	- \widehat{\mathbf{k}}^\top (K + T^{-1})^{-1} \widehat{\mathbf{k}} \nonumber \\
	&= k(\hat{x}, \hat{x})
	- \widehat{\mathbf{k}}^\top \widetilde{T}^{\frac{1}{2}}
	B^{-1} \widetilde{T}^{\frac{1}{2}} \widehat{\mathbf{k}},
\end{align}
where 
\begin{align}
	\widehat{\mathbf{k}} &=
	\begin{pmatrix}
		1 \\ \rho
	\end{pmatrix} \otimes
	\begin{pmatrix}
		k(\hat{x}, x_1) \\
		\vdots \\
		k(\hat{x}, x_n)
	\end{pmatrix}.
\end{align}

The predictive distribution for the target labels can also be calculated,
using the predictive distribution of the latent function value as follows:
\begin{align}
	p(\hat{y}=1| \hat{x}, X, \mathbf{y}, \mathbf{s})
	&= \Phi\left(\frac{\hat{\mu}}{\sqrt{1 + \hat{\sigma}^2}}\right).
\end{align}

\subsection{Marginal likelihood}

Usually, a covariance function
$k: \mathcal{X} \times \mathcal{X} \rightarrow \mathbb{R}$ has several hyperparameters.
For example, the radial basis function kernel $k_\text{RBF}(x, x')$
has a length scale parameter $l^2$:
\begin{align}
	k_\text{RBF}(x, x') = \exp\left(-\frac{\|x-x'\|^2}{2l^2}\right).
\end{align}
Additionally, there is a task similarity parameter $\rho$
that controls the importance of soft labels.
The hyperparameters including the kernel hyperparameters and the task similarity
are typically estimated by the empirical Bayes method,
which maximizes the marginal likelihood of the model.

The marginal likelihood of both target and source labels can be
approximated by using the site approximation.
Since our problem is a transfer learning problem,
our objective is to maximize the marginal likelihood
of target labels conditioned on source labels.
Although what we can compute in EP is
the joint marginal likelihood over target and source labels,
the conditional marginal likelihood can be computed
using Bayes' theorem as follows:
\begin{gather}
	Z_{X, \mathbf{y}|\mathbf{s}} := \log p(\mathbf{y} | \mathbf{s}, X)
	= \log \frac{p(\mathbf{y}, \mathbf{s} | X)}{p(\mathbf{s} | X)} \nonumber \\
	= \log p(\mathbf{y}, \mathbf{s} | X) - \log p(\mathbf{s} | X).
\end{gather}
The latter term $\log p(\mathbf{s}|X)$ is the log marginal likelihood over
soft labels.
Since the likelihood for soft labels is Gaussian,
the marginal likelihood can be calculated analytically.

\section{PAC-Bayes bound}\label{sec:bound}

In this section, we provide a theoretical analysis of our method based on
the probably approximately correct (PAC) Bayesian theorem~\cite{mcallester1999ML}.
The PAC-Bayesian theorem gives an upper bound of the generalization error
for randomized estimators.
We focus on the expected risk with the negative log likelihood loss
\begin{align}
\mathcal{L}_\mathcal{D}^{\ell_\text{nll}}(f_T)
= \expect_{(x, y)\sim \mathcal{D}} \ell_\text{nll}(f_T, x, y),
\end{align}
where $\mathcal{D}$ is a data generating distribution over the pair of
input $\mathcal{X}$ and output $\mathcal{Y} = \{+1, -1\}$ and
$\ell_\text{nll}$ is the negative log likelihood loss of the target task defined by
\begin{align}
\ell_\text{nll}(f_T, x, y) = -\log p(y|f_T(x)) = -\log \Phi(yf_T(x)).
\end{align}
The expected risk for our proposed method can be bounded by the log marginal likelihood as in
Theorem~\ref{thm:pac_slt}.

\begin{theorem}\label{thm:pac_slt}
	If the data distribution $\mathcal{D}$ over $\mathcal{X} \times \mathcal{Y}$
	and the prior distribution $P$ over
	a set of hypothesis $\mathcal{F} \subset \{f:\mathcal{X} \rightarrow \mathbb{R}\}$
	have the following sub-Gaussian property
	with a variance factor $\sigma_0^2 < \frac{1}{2}$:
	\begin{align}
		\forall \lambda \in \mathbb{R}:\;
		\log \expect_{f_T\sim P} \expect_{(x', y') \sim \mathcal{D}}
		\exp\left(\lambda y'f_T(x')\right)
		\le \frac{\lambda^2 \sigma_0^2}{2},
	\end{align}
	then the expectation of the expected risk over the posterior distribution
	$\widehat{Q}(f_T) = p(f_T|X, \mathbf{y}, \mathbf{s})$
		is bounded with probability at least $1-\delta$ as follows:
	\begin{align}
		\expect_{f_T\sim \widehat{Q}}
		\mathcal{L}_\mathcal{D}^{\ell_\textnormal{nll}}(f_T)
		&\le -\frac{1}{n}\log(\delta Z_{X, \mathbf{y}|\mathbf{s}})
		+ b_{\sigma_0^2},
	\end{align}
	where
	\begin{align}
		b_{\sigma_0^2} &= \inf_{a > c_{\sigma_0^2}}
		\left[\frac{1}{2}\log(2\pi(a+4)) \right.\nonumber \\
		&\qquad\qquad \left.-\frac{a}{2(a+4)}
		+ 4 \sigma_0^4 \left(\frac{a+5}{a+4}\right)^2\right], \\
		c_{\sigma_0^2} &= \frac{10\sigma_0^2-4}{1-2\sigma_0^2}.
	\end{align}
\end{theorem}

Theorem~\ref{thm:pac_slt} suggests that
the negative log marginal likelihood $-\log Z_{X, \mathbf{y}|\mathbf{s}}$
becomes an upper bound of the expected risk with additional constants
when $\sigma_0^2$ is fixed or bounded.
The value of $\sigma_0^2$ itself is generally hard to calculate analytically.
However, by choosing the covariance function in the Gaussian process prior
so that it has a small enough value,
the sub-Gaussian condition is expected to be satisfied.

\textbf{Proof of Theorem~\ref{thm:pac_slt}.}
We start with the previous bound~\cite{alquier2016JMLR, germain2016NIPS}.
The PAC-Bayes bound gives the relationship between
the expected risk
$\mathcal{L}_\mathcal{D}^\ell(f) = \expect_{(x,y)\sim \mathcal{D}}\ell(f, x, y)$ and
the empirical risk
$\widehat{\mathcal{L}}_{(X, \mathbf{y})}^\ell(f)
= \frac{1}{n} \sum_{i=1}^n \ell(f, x_i, y_i)$,
where
$\ell:\mathcal{F}\times\mathcal{X}\times\mathcal{Y}$ is a loss function over
a hypothesis set $\mathcal{F}\subset\{f:\mathcal{X}\rightarrow\mathbb{R}\}$,
input space $\mathcal{X}$ and output space $\mathcal{Y}$,
and $(X, \mathbf{y})\sim \mathcal{D}^n$.
The relationship between the expected and empirical risk is
represented by an inequality with the Kullback-Leibler divergence
from the posterior distribution $Q$ to the prior distribution $P$ over $\mathcal{F}$,
$\text{KL}[Q||P] = \expect_{f\sim Q} \left[\log\frac{Q(f)}{P(f)}\right]$.
The statement of the theorem is described as follows.

\begin{theorem}[\citealt{alquier2016JMLR}]\label{thm:alquier}
	Given a distribution $\mathcal{D}$ over $\mathcal{X} \times \mathcal{Y}$,
	a hypothesis set $\mathcal{F}$, a loss function
	$\ell: \mathcal{F}\times\mathcal{X}\times\mathcal{Y}\rightarrow\mathbb{R}$,
	a prior distribution $P$ over $\mathcal{F}$ and
	real numbers $\delta \in (0, 1]$ and $\lambda > 0$,
	the following inequality holds
	with probability at least $1 - \delta$
	over the choice of $(X, \mathbf{y}) \sim \mathcal{D}^n$:
	\begin{multline}
		\forall Q \;\text{on } \mathcal{F}: \quad
		\expect_{f\sim Q} \mathcal{L}_\mathcal{D}^\ell (f) \le
		\expect_{f\sim Q} \widehat{\mathcal{L}}_{X, \mathbf{y}}^\ell (f) \\
		+ \frac{1}{\lambda} \left[\textnormal{KL}[Q||P] + \log \frac{1}{\delta}
		+ \Psi_{\ell, P, \mathcal{D}}(\lambda, n)\right],
	\end{multline}
	where
	\begin{multline}
		\Psi_{\ell, P, \mathcal{D}}(\lambda, n)
		= \\ \log \expect_{f\sim P}\expect_{(X',\mathbf{y}')\sim \mathcal{D}^n}
		\exp\left[\lambda\left(
				\mathcal{L}_\mathcal{D}^\ell(f)
				-\widehat{\mathcal{L}}_{X', \mathbf{y}'}^\ell(f)
		\right)\right].
	\end{multline}
\end{theorem}

From Theorem~\ref{thm:alquier} and
the discussion by~\citet{germain2016NIPS},
we can relate the PAC-Bayesian theorem to the empirical Bayes method.
We consider the negative log likelihood loss
$\ell_\text{nll} (f, x, y) = -\log p(y|f(x))$ and
the optimal Gibbs posterior
\begin{align}
	\widehat{Q}(f) &= \frac{P(f)\exp(
	-n \widehat{\mathcal{L}}_{X, \mathbf{y}}^{\ell_\text{nll}}(f))}
	{Z_{X, \mathbf{y}}},
\end{align}
where
\begin{align}
	Z_{X, \mathbf{y}} &= \int P(f) \exp(
	-n\widehat{\mathcal{L}}_{X, \mathbf{y}}^{\ell_\text{nll}}(f)) \text{d}f
\end{align}
is the marginal likelihood of the training data $(X, \mathbf{y})$.
By substituting $n$ for $\lambda$,
the expectation of the expected risk is rewritten as follows:
\begin{align}
	\expect_{f\sim \widehat{Q}} \mathcal{L}_\mathcal{D}^{\ell_\text{nll}}(f)
	&\le -\frac{1}{n} \log (\delta Z_{X,\mathbf{y}})
	+ \frac{1}{n}\Psi_{\ell_\text{nll},P,\mathcal{D}}(n, n). \label{eq:before_bound}
\end{align}
In our method, the marginal likelihood of the target hard labels
are conditioned on the soft labels $p(\mathbf{y}| X, \mathbf{s})$.
This modification can be understood as modification of the
prior distribution, i.e.,\
\begin{align}
	P(f_T) = p(f_T|X, \mathbf{s}) = \mathcal{GP}(f_T | m_{X, \mathbf{s}}, k_{X}),
\end{align}
where
\begin{align}
	m_{X, \mathbf{s}}(x) &= \rho \mathbf{k}_X(x) K_X^{-1} \mathbf{s}, \\
	k_{X}(x, x') &= k(x, x') - \rho^2 \mathbf{k}_X(x) K_X^{-1} \mathbf{k}_X(x'), \\
	\mathbf{k}_X(x) &= (k(x, x_1), \ldots, k(x, x_n))^\top.
\end{align}
The marginal likelihood for our method is defined as follows:
\begin{align}
	Z_{X, \mathbf{y}|\mathbf{s}}
	&= \int p(f_T | X, \mathbf{s}) \prod_{i=1}^n p(y_i | f_T(x_i)) \text{d}f_T.
\end{align}
From the following lemma, the last term in \eqref{eq:before_bound}
can be bounded by a constant.

\textbf{Lemma 3}.
\begin{align}
	\Psi_{\ell_\text{nll},P,\mathcal{D}}(n, n) \le n b_{\sigma_0^2}.
\end{align}
The proof of Lemma 3 is provided in
\iflong
	Appendix~\ref{sec:proof}.
\else
	the supplementary material.
\fi

\begin{table*}[t!]
	\centering
	\caption{The data-generating processes for synthetic datasets where
		$d = 50$, $d^*=3$, $c=2$, $c^*=2$,
		$J = \{1, 2, 3\}$,
		$k_0(x, x') = 10 \exp \left(-\frac{\|x-x'\|^2}{2}\right)$,
		$k_0^*(x, x') = 10 \exp \left(-\frac{\|x-x'\|^2}{2}\right)$ and
		$\mathcal{J} = \{A \subset \{1, \ldots, d\} | |A| = d^*\}$.
		$\alpha \in \mathbb{R}^d$ and $\alpha^* \in \mathbb{R}^{d^*}$ are
		sampled from the standard normal distribution for each dataset.
		$\mathcal{U}(\mathcal{Z})$ denotes the uniform distribution
		over the set $\mathcal{Z}$.
		200 samples were generated for training and
		1000 samples were generated for testing.}\label{tab:generate}
	\makebox[\linewidth]{
	\begin{tabular}{ccc}
		\toprule
		Clean soft label & Clean feature & Relevant feature \\
		\midrule
		$x_i \sim \mathcal{N}(\boldsymbol{0}, I_d)$ &
		$x_i^* \sim \mathcal{N}(\boldsymbol{0}, I_d)$ &
		$x_i \sim \mathcal{N}(\boldsymbol{0}, I_d)$ \\
		$x_i^* \leftarrow \alpha^\top x_i$ &
		$\epsilon_i \sim \mathcal{N}(\boldsymbol{0}, I_d)$ &
		$x_i^* \leftarrow x_{i, J}$ \\
		$\epsilon_i \sim \mathcal{N}(0, 1)$ &
		$x_i \leftarrow x_i^* + \epsilon_i$ &
		$y_i \leftarrow \text{sign}({\alpha^*}^\top x_i^*)$ \\
		$y_i \leftarrow \text{sign}(x_i^* + \epsilon)$ &
		$y_i \leftarrow \text{sign}(\alpha^\top x_i^*)$ & \\
		\midrule
		\midrule
		Independent feature & Latent GP & Noise variance \\
		\midrule
		$x_i \sim \mathcal{N}(\boldsymbol{0}, I_d)$ &
		$x_i \sim \mathcal{U}([0, 10]^c)$ &
		$x_i \sim \mathcal{U}([0, 10]^c)$,
		$x_i^* \sim \mathcal{U}([0, 10]^{c^*})$ \\
		$J_i \sim \mathcal{U}(\mathcal{J})$ &
		$K_0 = {\left(k_0(x_i, x_j)\right)}_{i,j}$ &
		$K_0 = {\left(k_0(x_i, x_j)\right)}_{i,j}$,
		$K_0^* = \left(k_0^*(x_i^*, x_j^*)\right)_{i,j}$ \\
		$x_i^* \leftarrow x_{i, J_i}$ & 
		$(x_1^*, \ldots,  x_n^*)^\top \sim \mathcal{N}(\boldsymbol{0}, K_0)$ &
		$(f_1, \cdots, f_n)^\top \sim \mathcal{N}(\boldsymbol{0}, K_0)$ \\
		$y_i \leftarrow \text{sign}({\alpha^*}^\top x_i^*)$ &
		$\epsilon_i \sim \mathcal{N}(0, 1)$ &
		$(g_1, \cdots, g_n)^\top \sim \mathcal{N}(\boldsymbol{0}, K_0^*)$ \\
		& $y_i \leftarrow \text{sign}(x_i^* + \epsilon_i)$ &
		$y_i \leftarrow \text{sign}(f_i + \epsilon_i)$, 
		$\epsilon_i \sim \mathcal{N}(0, \exp(g_i))$ \\
		\bottomrule
\end{tabular}}
	\vskip -0.5cm
\end{table*}
\begin{table*}[t!]
\caption{Results of synthetic datasets. Each experiment was repeated 100 times.
The mean and standard deviation (enclosed in parentheses) of accuracy
are displayed. The best accuracy for each dataset except the rightmost column
is highlighted in \textbf{boldface}.
The rightmost column is the test accuracy of GPC
where privileged information is given as input data
for both training and test dataset.}\label{tab:synthetic}
\centering
\makebox[\linewidth]{
\begin{tabular}{cccc|c}
	\toprule
	Datasets & GPC & GPC+ & SLT-GP & GPC (reference) \\
	\midrule
	Train & $X$ & $(X, X^*)$ & $(X, X^*)$ & $X^*$ \\
	Test & $X$ & $X$ & $X$ & $X^*$ \\
	\midrule
	Clean soft label    & 87.89 (2.39) & 88.21 (1.57)          & \textbf{95.27 (0.89)} & 95.41 (0.82) \\
	Clean feature       & 65.40 (2.31) & 68.89 (2.01)          & \textbf{69.99 (1.65)} & 89.72 (1.38) \\
	Relevant feature    & 89.85 (1.82) & 89.88 (1.85)          & \textbf{98.92 (0.57)} & 99.09 (0.58) \\
	Independent feature & 50.68 (1.68) & 50.81 (1.80)          & \textbf{50.95 (1.78)} & 99.01 (0.58) \\
	Latent GP           & 82.20 (2.83) & 80.05 (9.14)          & \textbf{86.75 (2.35)} & 89.97 (1.77) \\
	Noise variance      & 77.36 (6.05) & \textbf{78.85 (6.54)} & 77.90 (5.97)          & 55.34 (5.09) \\
	\bottomrule
\end{tabular}}
	\vskip -0.5cm
\end{table*}

\section{Experiments}

In this section, we present the results of experiments on
synthetic and real-world datasets.
In the experiments on synthetic datasets in
Section~\ref{subsec:cleanlabel}--\ref{subsec:noise_variance},
our proposed method (Soft-Label Transferred Gaussian Process classification, SLT-GP)
was compared with three methods---Gaussian process classification (GPC)
with only privileged information (treated as input data),
standard GPC without privileged information, and GPC+.
The performance of GPC with only privileged information is
calculated by privileged information of test data,
which is not actually available in our problem setting
but its results are useful to understand the improvement of the algorithm.
The experimental settings
of Section~\ref{subsec:cleanlabel}--\ref{subsec:independent}
were originally presented in the previous work~\cite{lopez2016ICLR}.
The data-generating processes for the synthetic datasets 
in Section~\ref{subsec:cleanlabel}--\ref{subsec:noise_variance} are summarized
in Table~\ref{tab:generate} and their results are given in Table~\ref{tab:synthetic}.
In Section~\ref{subsec:flowcap},
the performance of our proposed SLT-GP method and GPC+ is compared
by using real-world datasets.
Finally, we investigate the change in the task similarity $\rho$ estimated by
empirical Bayes method compared with the optimal one in Section~\ref{subsec:rho}.
The hyperparameters in the covariance function of each method and
the task similarity are estimated by the empirical Bayes method for all experiments.

\subsection{Clean soft labels as privileged information}\label{subsec:cleanlabel}

In this setting, the true distance
(including the sign corresponding to the target class)
from the decision boundary is given as privileged information.
The data-generating process is presented in the Table~\ref{tab:generate} as
``clean soft label.''
We used the linear kernel,
$k_\text{linear}(x, x') = \sigma^2 x^\top x'$,
as the covariance function of the latent function over input for GPC, SLT-GP and GPC+.
The Gaussian radial basis function
$k_\text{RBF}(x, x') = \exp\left(-\frac{\|x-x'\|^2}{2l^2}\right)$
was used as the covariance function of the latent function over privileged information for GPC+
because it is assumed that the relationship between
the privileged information and
the latent function over the privileged information is not linear.
The row ``clean soft label'' in Table~\ref{tab:synthetic} shows
the result of this experiment.
This result shows that our method outperforms the baselines
because the privileged information and 
the soft labels generated from the privileged information
have enough information to classify each sample.

\subsection{Clean features as privileged information}\label{subsec:cleanfeature}

In this setting, each input feature is contaminated and
the feature before contamination is given as privileged information.
The data-generating process is presented in Table~\ref{tab:generate} as
``clean feature.''
The covariance functions used in each method were the same as the experiment
in Section~\ref{subsec:cleanlabel}.
The results of this experiment are shown in Table~\ref{tab:synthetic}
as the dataset ``clean feature.''
In this dataset, the relationship between the input data and
the privileged information is weak because
the input is contaminated by noise.
Therefore, the privileged information is not so useful
and there are few differences between our method and the baselines.

\subsection{Relevant features as privileged information}\label{subsec:relevant}

In this setting, only a subset of all the features is relevant to the output label.
The relevant features were given as privileged information.
The data-generating process is presented in Table~\ref{tab:generate} as
``relevant feature.''
We used the same covariance functions as the ones described in Section~\ref{subsec:cleanlabel}
for each method.
The results of this experiment are shown in Table~\ref{tab:synthetic}
as the dataset ``relevant feature.''
Since only the subset of all the features were given as
privileged information, unnecessary information is eliminated from
the input data. Therefore, privileged information improved the performance
of our method. However, the improvement for GPC+ was negligible.

\subsection{Independent relevant features as privileged information}\label{subsec:independent}

In this setting, only the subset of all the features is
relevant as the previous dataset ``relevant feature.''
But relevant features are independent for each data example.
The data-generating process is presented in Table~\ref{tab:generate} as
``independent feature.''
We used the same covariance functions as the ones described in Section~\ref{subsec:cleanlabel}
for each method.
The results of this experiment are shown in Table~\ref{tab:synthetic}
as the dataset ``independent feature.''
In this dataset,
though only the relevant features were given as privileged information,
the privileged information did not have the information about which features are
relevant to each output. Thus, the privileged information rarely improved the
performance over the standard GPC in both SLT-GP and GPC+.

\subsection{Latent Gaussian process as privileged information}\label{subsec:latent_gp}

In this setting, the data was generated from a latent Gaussian process.
The output was the sign of the value of the latent function,
while the privileged information was the value of the latent function itself.
This setting is a non-linear version of Section~\ref{subsec:cleanlabel}.
The data-generating process is presented in Table~\ref{tab:generate} as
``latent GP.''
We used the Gaussian radial basis function $k_\text{RBF}$ as
the covariance function in all methods.
The results of this experiment are shown in Table~\ref{tab:synthetic}
as the dataset ``latent GP.''
This result shows that our method outperforms other baselines
even with the non-linear dataset.

\subsection{Noise variance as privileged information}\label{subsec:noise_variance}

In this setting, the noise variance of the latent function depends on
the privileged information.
This dataset simulates the assumption of the data generation process of GPC+,
which is presented in Table~\ref{tab:generate} as ``noise variance.''
We used the Gaussian radial basis function $k_\text{RBF}$ as
the covariance function in all methods.
The results of this experiment are shown in Table~\ref{tab:synthetic}
as the dataset ``noise variance.''
In this dataset, the accuracy of GPC+ is slightly higher than our proposed method.
However, the improvement is marginal.

\begin{figure}
	\centering
	\includegraphics[width=1.0\columnwidth]{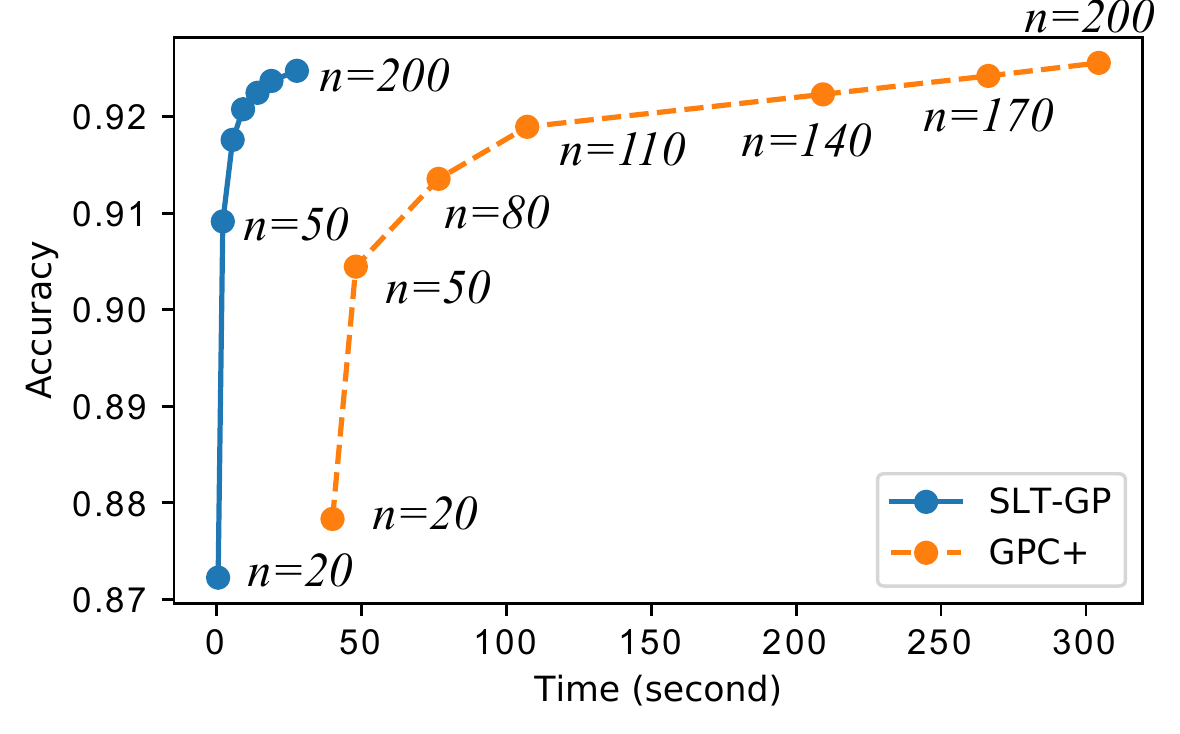}
	\vskip -0.5cm
	\caption{Computation time versus accuracy.
	Dataset size was changed from 20 to 200,
	repeating 100 times for each dataset size.
	The horizontal axis represents computation time in seconds and
	the vertical axis represents the accuracy.}\label{fig:flowcap}
	\vskip -0.5cm
\end{figure}

\subsection{Real-world dataset (FlowCAP)}\label{subsec:flowcap}

We consider analyzing real-world cell data measured by a flow cytometer.
As a possible case, we can consider computer-aided diagnoses based on flow cytometry
in the following situation.
In large hospitals, there are highly functional flow cytometers,
while many small hospitals have only limited functional flow cytometers.
In such cases, diagnoses in small hospitals might be improved
by using the datasets obtained by highly functional flow cytometers as
privileged information even though limited input features are available for diagnoses.

In this experiment, we used the FlowCAP dataset~\cite{dundar2014KDD}.
The dataset has three types of cells, one for a class of normal cells and two
for abnormal cells.
Each cell has a six-dimensional feature vector:
two light scatter characteristics and four fluorescence intensity values.
In our preliminary experiment, two of the fluorescence intensity values were
good at separating the normal and abnormal classes.
Therefore, we used these two features as privileged information and
remaining four features as input data.
$200$ examples were sampled for training and $1000$ examples for testing.
The numbers of positive examples (abnormal cells) and negative examples (normal cells)
are the same in the training data and test data, respectively.
The covariance functions used for all four methods were the Gaussian radial basis function kernels
$k_\text{RBF}$.

The classification accuracy and computation time are plotted
in Figure~\ref{fig:flowcap},
where the number of training data is increased from 20 to 200.
As the figure shows, our proposed method achieves almost the same accuracy
as GPC+ with much lower computation time.

\subsection{On estimation of task similarity}\label{subsec:rho}

\begin{figure}
	\centering
	\includegraphics[width=1.0\columnwidth]{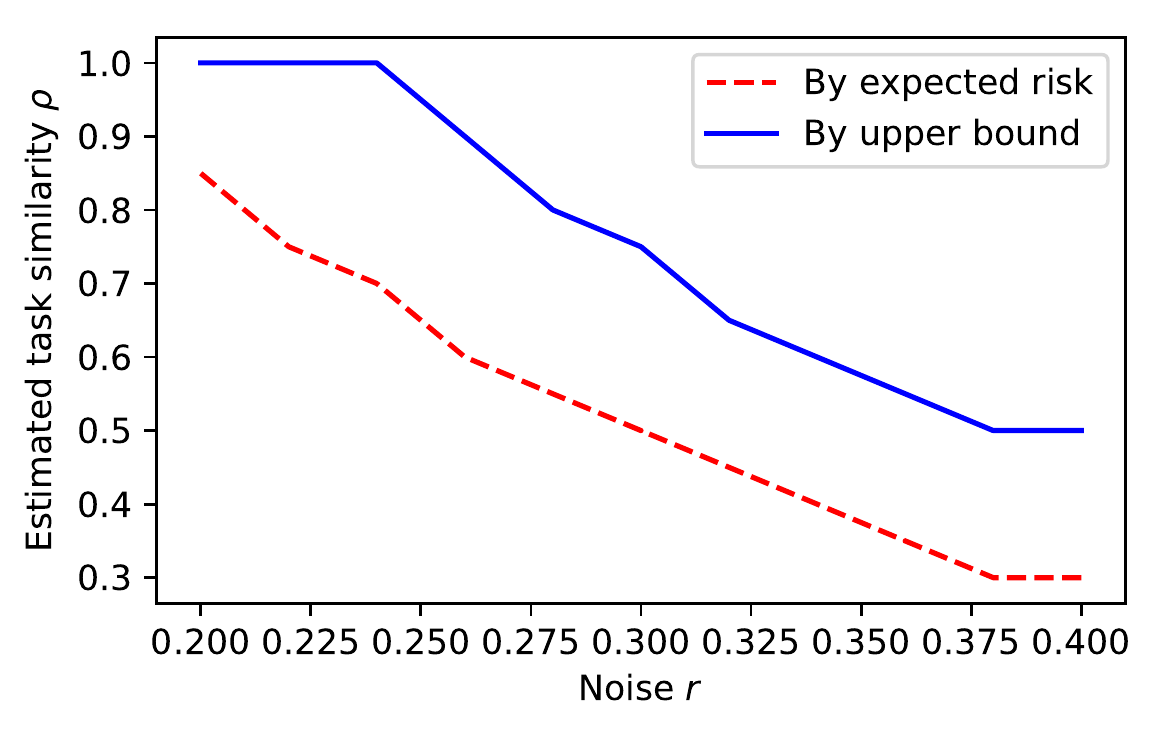}
	\vskip -0.5cm
	\caption{Changes in optimal $\rho$ in terms of two objectives,
	the upper bound of expected risk (blue solid line) and
	the expected risk calculated with test data (red dashed line).}
	\label{fig:rho_change}
	\vskip -0.4cm
\end{figure}

Finally, we numerically investigate the change in the task similarity $\rho$
in the case where the privileged information is contaminated.
Also, the relationship
between the true expected risk and its upper bound
(used in the empirical Bayes method) is explained by estimating $\rho$.

We synthetically generated data as follows:
\begin{gather*}
	x_i \sim \mathcal{U}([0, 10]^2) \quad (i=1,\ldots, n), \\
	(f_1^*, \ldots,  f_n^*)^\top \sim \mathcal{N}(\boldsymbol{0}, 10 K_0), \\
	g_i^* \sim \mathcal{N}(0, 10) \quad (i=1,\ldots, n), \\
	\epsilon_i \sim \mathcal{N}(0, 1) \quad (i=1,\ldots, n), \\
	x_i^* \leftarrow (1-r) f_i^* + r g_i^* \quad (i=1,\ldots, n), \\
	y_i \leftarrow \text{sign}(f_i^* + \epsilon_i) \quad (i=1,\ldots, n),
\end{gather*}
where $r \in [0, 1]$ is a parameter controlling the noise in privileged information,
\begin{align}
K_0 &= \left(k_0(x_i, x_j)\right)_{i, j}, \\
k_0(x, x') &= \exp\left(-\frac{\|x-x'\|^2}{2}\right).
\end{align}
We generated $200$ samples as training data and
observed the changes in $\rho$ chosen by two objectives,
which was repeated $100$ times for different value of $r$.
We compared $\rho$ calculated by minimizing our upper bound of the expected risk
(equivalent to the empirical Bayes method) with $\rho$
calculated by minimizing the true expected risk calculated with $1000$ test data.
We used the Gaussian radial basis function with the fixed variance amplitude
$k(x, x') = \frac{1}{4} \exp\left(-\frac{\|x-x'\|^2}{2l^2}\right)$,
where $l^2$ is a length-scale parameter optimized by the empirical Bayes method.

The result is shown in Figure~\ref{fig:rho_change}.
The changes in $\rho$ have the same tendency between the two objectives and
$\rho$ becomes smaller as the noise rate becomes larger.
However, hyperparameter tuning based on the upper bound
systematically yielded slightly larger values of $\rho$ than
the optimal ones in terms of the expected risk.
The difference in the value of $\rho$ might have been resulted
from the constant term in our bound.
The difference would have been reduced
if we can derive a computable tighter upper bound,
which is left as future work.

\section{Conclusion}

Our contributions are summarized as follows.
\begin{enumerate*}[label={(\Alph*)}]
\item We proposed a novel classification method
	with privileged information (SLT-GP).
\item We analyzed our proposed method by the PAC-Bayesian theorem.
\item We empirically compared our proposed method and the previous methods
	based on Gaussian processes.
\end{enumerate*}

Our proposed method incorporates privileged information as
soft labels that contain the information about the confidence
belonging to each class and transfers the information to the target task
predicting the hard labels.
This formulation includes a hyperparameter corresponding to ``task similarity,''
which controls how much information should be transferred from 
the soft-labeling task to the hard-labeling task.
The task similarity parameter can be determined by the empirical Bayes method
and becomes an interpretable index for the effect of the privileged information.

There are several possible directions to extend this work.
In this work, we have only considered binary classification tasks.
However, it would be interesting to extend our work to multi-class classification tasks
because soft labels of multi-class classification tend to have
much more information about the privileged information.
Another area of research is devising a different hyperparameter optimization scheme
based on the PAC-Bayesian bounds from the empirical Bayes method.
In our numerical simulation, there was a gap between the expected risk and
the upper bound.
The gap and the expected risk would be reduced
if we could derive a tighter upper bound and directly minimize it.

\section*{Acknowledgements}

MS was supported by JST CREST JPMJCR1403.
IS was supported by KAKEN 17H04693.

\bibliographystyle{plain}
\bibliography{references}

\appendix

\appendix

\section{Proof of Lemma 3}\label{sec:proof}

In the discussion by \citet{germain2016NIPS},
the following assumption is considered implicitly,
\begin{multline}
	\expect_{f\sim P} \expect_{(X', \mathbf{y}') \sim \mathcal{D}^n}
	\exp \left[\sum_{i=1}^n V_i(f)\right] \\
	= \expect_{f_1, \ldots, f_n \sim P} \expect_{(X', \mathbf{y}') \sim \mathcal{D}^n}
	\exp \left[\sum_{i=1}^n V_i(f_i)\right],
\end{multline}
where
\begin{align}
	V_i(f) &= \mathcal{L}_\mathcal{D}^{\ell_\text{nll}}(f)
	- \ell_\text{nll}(f, x_i', y_i').
\end{align}
By using the same assumption,
the term $\Psi_{\ell_\text{nll}, P, \mathcal{D}}(n, n)$
can be modified as follows:
\begin{align}
	\Psi_{\ell_\text{nll}, P, \mathcal{D}(n, n)}
	&= \log \expect_{f\sim P} \expect_{(X', \mathbf{y}') \sim \mathcal{D}^n} \\
	&\qquad \exp \left[
		n \left(
			\mathcal{L}_\mathcal{D}^{\ell_\text{nll}}(f)
			-\widehat{\mathcal{L}}_{X', \mathbf{y}'}^{\ell_\text{nll}}(f)
		\right)
	\right] \nonumber \\
	&= \log \expect_{f\sim P} \expect_{(X', \mathbf{y}')\sim\mathcal{D}^n}
	\exp \left[
		\sum_{i=1}^n V_i(f)
	\right] \nonumber \\
	&= \log \expect_{f_1, \ldots, f_n \sim P}
	\expect_{(X', \mathbf{y}')\sim\mathcal{D}^n}
	\exp \left[
		\sum_{i=1}^n V_i(f_i)
	\right] \nonumber \\
	&= \sum_{i=1}^n \log \expect_{f\sim P} \expect_{(x_i', y_i')\sim \mathcal{D}}
	\exp V_i(f).
	\label{eq:psi_mod}
\end{align}
We assume that $y'f(x')$ over the prior distribution $P$ is
sub-Gaussian with the variance factor $\sigma_0^2$, i.e.\
\begin{align}
	\log \expect_{f\sim P} \expect_{(x', y') \sim \mathcal{D}}
	\exp \lambda y'f(x')
	\le \frac{\lambda^2 \sigma_0^2}{2} \quad(\lambda \in \mathbb{R}).
\end{align}
From the fact about the cumulative distribution function
of normal distribution~\cite{komatu1955},
\begin{align}
	\Phi(z) &> \frac{2\phi(z)}{\sqrt{z^2+4}-z} \quad (z\le 0),
\end{align}
the inverse of cumulative distribution function can be evaluated for any $a > -4$,
\begin{align}
	\Phi(z)^{-1} &< \frac{\sqrt{z^2+4}-z}{2\phi(z)} \nonumber \\
	&= \frac{\sqrt{2\pi}(\sqrt{z^2+4}-z)}{2}\exp\left(\frac{z^2}{2}\right) \nonumber \\
	&\le \sqrt{2\pi} \exp\left(\frac{1}{2}z^2 + \frac{1}{2}\log (z^2+4)\right) \nonumber \\
	&\le \sqrt{2\pi} \exp\left(\frac{1}{2}z^2
		+ \frac{1}{2}\left(\frac{z^2-a}{a+4}+\log(a+4)\right)\right) \nonumber \\
	&= \sqrt{2\pi(a+4)} \exp\left(\frac{a+5}{2(a+4)}z^2
	- \frac{a}{2(a+4)}\right). \label{eq:phi_inv_ub}
\end{align}
Note that the last inequality holds for any $z \in \mathbb{R}$.
If $z$ is sub-Gaussian with a variance factor $\sigma_z^2$ such that
$\frac{a+5}{2(a+4)} < \frac{1}{4\sigma_z^2}$,
the following inequality holds~\cite{honorio2014AISTATS},
\begin{align}
	\log \expect_z \Phi(z)^{-1}
	&\le \log \expect_z \left[\sqrt{2\pi(a+4)}\right. \nonumber \\
	&\qquad \qquad \left.\exp\left(
	\frac{a+5}{2(a+4)}z^2 - \frac{a}{2(a+4)}\right)\right] \nonumber \\
	&= \frac{1}{2}\log(2\pi(a+4)) - \frac{a}{2(a+4)} \nonumber \\
	&\quad + \log \expect_z \exp\left(\frac{a+5}{2(a+4)}z^2\right) \nonumber \\
	&\le \frac{1}{2}\log(2\pi(a+4)) - \frac{a}{2(a+4)} \nonumber \\
	&\quad + 4 \left(\frac{a+5}{a+4}\right)^2 \sigma_z^4.
\end{align}
\begin{samepage}
Each term in~\eqref{eq:psi_mod} is evaluated as follows:
\begin{multline}
	\log \expect_{f\sim P} \expect_{(x_i', y_i')\sim \mathcal{D}} \exp V_i(f) \\
	\begin{aligned}[b]
	&\le \log \expect_{f\sim P}
	\exp \mathcal{L}_\mathcal{D}^{\ell_\text{nll}}(f) \\
	&\le \log \expect_{f\sim P} \expect_{(x',y')\sim \mathcal{D}}
	\exp \ell_\text{nll}(f, x', y') \\
	&= \log \expect_{f\sim P} \expect_{(x',y')\sim \mathcal{D}}
	\exp (-\log \Phi(y' f(x'))) \\
	&= \log \expect_{f\sim P} \expect_{(x',y')\sim \mathcal{D}}
	\Phi(y'f(x'))^{-1} \\
	&\le \frac{1}{2}\log(2\pi(a+4)) - \frac{a}{2(a+4)} \\
	&\quad + 4 \left(\frac{a+5}{a+4}\right)^2 \sigma_0^4.
	\end{aligned}
\end{multline}
\end{samepage}
Note that the last inequality~\eqref{eq:phi_inv_ub} holds for any $z \in \mathbb{R}$.

The condition that the above inequality holds is
\begin{multline}
	\frac{a+5}{2(a+4)} < \frac{1}{4\sigma_0^2}
	\iff
		a > \frac{10\sigma_0^4 - 4}{1-2\sigma_0^2}.
\end{multline}

\end{document}